# A Novel Model for Driver Lane Change Prediction in Cooperative Adaptive Cruise Control Systems

Armin Nejadhossein Qasemabadi[1], Saeed Mozaffari[2], Mahdi Rezaei[3], Majid Ahmadi[1], Shahpour Alirezaee[2]
[1] Electrical and Computer Engineering Department. University of Windsor, Windsor, Canada
[2] Mechanical, Automotive, and Material Engineering Department. University of Windsor, Windsor, Canada
{nejadho, saeed.mozaffari, m.ahmadi, s.alirezaee } @uwindsor.ca
[3]Institute for Transport Studies, University of Leeds, UK.  {m.rezaei@leeds.ac.uk}

*Abstract*— **Accurate lane change prediction can reduce potential accidents and contribute to higher road safety. Adaptive cruise control (ACC), lane departure avoidance (LDA), and lane keeping assistance (LKA) are some conventional modules in advanced driver assistance systems (ADAS). Thanks to vehicle-to-vehicle communication (V2V), vehicles can share traffic information with surrounding vehicles, enabling cooperative adaptive cruise control (CACC). While ACC relies on the vehicle's sensors to obtain the position and velocity of the leading vehicle, CACC also has access to the acceleration of multiple vehicles through V2V communication. This paper compares the type of information (position, velocity, acceleration) and the number of surrounding vehicles for driver lane change prediction. We trained an LSTM (Long Short-Term Memory) on the HighD dataset to predict lane change intention. Results indicate a significant improvement in accuracy with an increase in the number of surrounding vehicles and the information received from them. Specifically, the proposed model can predict the ego vehicle lane change with 59.15% and 92.43% accuracy in ACC and CACC scenarios, respectively.**

## I. Introduction

The number of vehicles on roads and highways has soared in recent years and we are witnessing more traffic congestion and vehicle accidents on the streets. To address these issues, car manufacturers have been developing advanced driver assistance systems (ADAS) [1]. Currently, ADAS has various modules for safe and convenient driving including collision avoidance system (CA), adaptive cruise control (ACC), and lane keeping assistance (LKA). The next generation of ADAS and automated vehicles will rely on advanced sensor technology and leverage artificial intelligence to predict the drivers' behavior and readiness and take appropriate measures in advance, to avoid accidents [2].

Lane changing is one of the most important behaviors of drivers as it is the main cause of vehicle collisions. Accurate lane change (LC) prediction will lead to improved vehicle safety and passengers' comfort. Lane change prediction is a subset of trajectory prediction in which spatial coordinates of vehicles are predicted in the future time. Unlike trajectory prediction, lane change prediction aims to predict if the driver drives away from the current lane and merges into adjacent lanes or keeps the current lane for driving. In this context, the vehicle equipped with ADAS and automated driving functions is called the ego vehicle  and other vehicles around the ego vehicle are referred to as surrounding vehicles. Lane change prediction can be divided into two main groups: driver lane change prediction [3] and surrounding vehicles' lane change prediction [4]. In other words, driver lane change prediction aims to predict the ego vehicle change lane, while surrounding vehicles' lane change prediction tries to forecast when another vehicle tries to cut-in in front of the ego vehicle from adjacent lanes.

There are several methods commonly used for lane change prediction in autonomous driving systems, including rule-based methods [5], machine learning-based methods [6], and sensor fusion-based methods [7]. Rule-based methods rely on speed difference and spacing between the vehicle and surrounding traffic to predict lane changes, while machine learning-based methods utilize trajectories of the surrounding vehicles to make predictions. Sensor fusion-based methods combine data from multiple sensors, such as cameras, lidar, and radar, or multiple vehicles information obtained through V2V to predict lane changes.

Agent-based methods, game theory and mixed logic programming have been used for rule-based lane change prediction [8]. Machine learning-based methods include Bayes classifier, support vector machine, hidden Markov model, or artificial neural network and deep learning algorithms [9]. Vehicle motion parameters such as steering wheel angle, driver's parameters like eye movement and head rotation, and surrounding vehicles information such as location, speed, and acceleration are combined in sensor fusion-based methods [10].

This paper focuses on the driver lane change prediction scenario and aims to explore the impact of the ego-vehicle's status (location, speed, acceleration) and the number of surrounding vehicles (ACC and CACC systems) on the lane change prediction accuracy. We used multiple long short-term memory (multi-LSTM) deep models which were trained and evaluated on a real traffic data set (HighD) [11].

## II. ACC and CACC Systems

According to the SAE level 3 (L3) autonomy, lane-changing algorithms are the basis of ACC systems in which the vehicle is capable of changing lanes under a human driver's supervision.

ACC systems typically use one or more sensors, such as radar, lidar, or cameras, to detect the distance and speed of the vehicle in front. Utilizing this information, the ego vehicle can follow the leading vehicle at a safe distance. However, the performance of the ACC systems is limited to the on-board sensors' range of approximately 150 meters and a field of view of approximately 20 degrees. Therefore, the CACC systems have emerged to supplant onboard sensors with vehicular communication to exchange information between the vehicles [12]. Unlike the ACC systems that only rely on distance and velocity measurements, the CACC can use extra information from adjacent vehicles such as their acceleration profile [13]. Therefore, the type of information and number of surrounding vehicles are different in ACC and CACC systems. In this paper, we assume that the ACC system can measure the position and velocity (two parameters) of lead vehicles in the current lane, left lane, and right lane (3 vehicles). In the CACC system, on the other hand, we assume that the position, velocity, and acceleration (three parameters) of lead and lag vehicles in the current lane, left lane, and right lane (6 vehicles), as well as adjacent vehicles in the left and right lanes (2 vehicles), are available.

## III. DATASET

In this paper, we used *HighD* which is a large-scale dataset containing high-resolution videos recorded by a drone from German highways [11]. The dataset contains the trajectories of more than 110,000 vehicles recorded at six different locations.

For lane change prediction specifically, the HighD dataset includes 5,600 complete lane changes performed by the drivers, as well as data on the surrounding vehicles and the driving environment. Compared to other datasets used for lane change prediction, this dataset has a larger size from the lane change point of view. For example, the number of lane changes in the HighD dataset is two times as much as NGSIM [14]. This is mainly due to a lower average traffic density and the larger number of lanes result. The data set has metadata which provides valuable information for lane change prediction such as the assigned ID to each vehicle, its (x,y) position, lateral/longitudinal velocity and acceleration of the vehicle, lane ID, as well as IDs of eight surrounding vehicles. Figure 1 shows the location of the ego vehicle, preceding/following vehicles (PV, FV) which are in the same lane with the ego vehicle, left preceding/ alongside/following (LP, LA, LF) vehicles which are in the adjacent lane on the left as well as right preceding/ alongside/following (RP, RA, RF) vehicles which are in the adjacent lane on the right.

## IV. PROPOSED METHOD

In the proposed LC prediction method, a LC commences when the vehicle's lane ID changes. After finding the vehicles with LC behavior and extracting the required information we train an LSTM to predict lane changing (LC) and lane keeping (LK) actions.

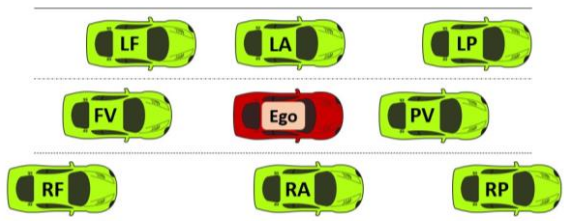

Figure 1.     Ego vehicle and its surrounding vehicles.

### A. Variables

To train and test the LSTM, first, we selected the vehicles with LC. Then, the corresponding LC frame is detected. After finding the LC frame ($f_{lc}$), we select *n* frames before the event and use [$f_{lc}$ – *n*, $f_{lc}$] frames as our training set. In other words, parameter *n* indicates the time length of the data set. Similarly, *n* frames will be used for training the LSTM for LK action. Finally, surrounding vehicles' parameters were extracted from the *HighD* data set which include relative distance, relative speed, and relative acceleration between surrounding vehicles and the ego vehicle. We will investigate the effect of these parameters and the number of surrounding vehicles on the LC behavior. Table 1 shows the behavior of vehicle number 48 which was in the 3$^{rd}$ lane from frame 1137 to frame 1147. At frame 1148, the vehicle moved to the 2$^{nd}$ lane. The ego vehicle was surrounded by vehicles number 45, 46, and 49. The ID value is set to 0, if no vehicle exists in the corresponding location.

The input vector of the LSTM model depends on the vehicle's information, and the type of surrounding vehicles:

$$x_t = [dp(i), dv(i), da(i)] \quad (1)$$

where *i* shows the type of vehicle which iterates over {LA, LP, PV, RP, RA, RF, FV, LF}. Parameters *dp(i)*, *dv(i)* and *da(i)* are Manhattan distances between position, velocity and acceleration of the ego vehicle and the *i*$^{th}$ vehicle, respectively. The length of the input vector also depends on the number of frames before LC.

$$X_t = [x_{t-n}, ..., x_{t-1}, x_t] \quad (2)$$

where *t* is equal to $f_{lc}$ for the car with LC and *n* is the time length.

### B. LSTM Model

LSTM stands for Long Short-Term Memory, which is a variant of Recurrent Neural Networks (RNN). As shown in Figure 2, our proposed multi-LSTM network consists of two LSTM layers, each consisting of several LSTM cells. The output of the second layer goes through a fully connected layer (FC) with 32 neurons to predict the binary value of 0 or 1, representing LK and LC actions respectively. Table 2 shows other parameters of our LSTM model.

TABLE I.     TIME-DEPENDENT VALUES FOR ONE SAMPLE VEHICLE WITH A LANE CHANGE.

| frame | id | x | y | xVel | yVel | xAcc | yAcc | PVId | FVId | LPId | LAId | LFId | RPId | RAId | RFId | laneId |
|---|---|---|---|---|---|---|---|---|---|---|---|---|---|---|---|---|
| 1137 | 48 | 166.64 | 12.11 | -33.64 | -1.17 | 0.44 | -0.21 | 46 | 49 | 0 | 0 | 0 | 0 | 0 | 45 | 3 |
| 1138 | 48 | 165.3 | 12.06 | -33.62 | -1.18 | 0.44 | -0.19 | 46 | 49 | 0 | 0 | 0 | 0 | 0 | 45 | 3 |
| 1139 | 48 | 163.96 | 12.01 | -33.6 | -1.19 | 0.44 | -0.17 | 46 | 49 | 0 | 0 | 0 | 0 | 0 | 45 | 3 |
| 1140 | 48 | 162.61 | 11.96 | -33.58 | -1.2 | 0.45 | -0.15 | 46 | 49 | 0 | 0 | 0 | 0 | 0 | 45 | 3 |
| 1141 | 48 | 161.26 | 11.91 | -33.57 | -1.21 | 0.45 | -0.13 | 46 | 49 | 0 | 0 | 0 | 0 | 0 | 45 | 3 |
| 1142 | 48 | 159.91 | 11.86 | -33.55 | -1.22 | 0.45 | -0.11 | 46 | 49 | 0 | 0 | 0 | 0 | 0 | 45 | 3 |
| 1143 | 48 | 158.57 | 11.81 | -33.53 | -1.22 | 0.45 | -0.08 | 46 | 49 | 0 | 0 | 0 | 0 | 0 | 45 | 3 |
| 1144 | 48 | 157.23 | 11.76 | -33.51 | -1.23 | 0.46 | -0.06 | 46 | 49 | 0 | 0 | 0 | 0 | 0 | 45 | 3 |
| 1145 | 48 | 155.89 | 11.71 | -33.49 | -1.23 | 0.46 | -0.04 | 46 | 49 | 0 | 0 | 0 | 0 | 0 | 45 | 3 |
| 1146 | 48 | 154.54 | 11.65 | -33.48 | -1.23 | 0.46 | -0.02 | 46 | 49 | 0 | 0 | 0 | 0 | 0 | 45 | 3 |
| 1147 | 48 | 153.19 | 11.6 | -33.46 | -1.23 | 0.46 | 0 | 46 | 49 | 0 | 0 | 0 | 0 | 0 | 45 | 3 |
| 1148 | 48 | 151.85 | 11.55 | -33.44 | -1.23 | 0.46 | 0.02 | 0 | 45 | 46 | 0 | 49 | 0 | 0 | 0 | 2 |
| 1149 | 48 | 150.51 | 11.5 | -33.42 | -1.23 | 0.47 | 0.04 | 0 | 45 | 46 | 0 | 49 | 0 | 0 | 0 | 2 |
| 1150 | 48 | 149.18 | 11.45 | -33.4 | -1.22 | 0.47 | 0.06 | 0 | 45 | 46 | 0 | 49 | 0 | 0 | 0 | 2 |
| 1151 | 48 | 147.85 | 11.4 | -33.38 | -1.22 | 0.47 | 0.08 | 0 | 45 | 46 | 0 | 49 | 0 | 0 | 0 | 2 |
| 1152 | 48 | 146.53 | 11.35 | -33.36 | -1.21 | 0.47 | 0.1 | 0 | 45 | 46 | 0 | 49 | 0 | 0 | 0 | 2 |
| 1153 | 48 | 145.2 | 11.3 | -33.34 | -1.21 | 0.47 | 0.12 | 0 | 45 | 46 | 0 | 49 | 0 | 0 | 0 | 2 |

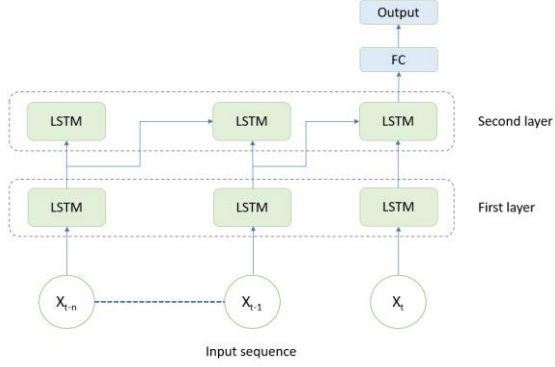

Figure 2. The architecture of the proposed LSTM network.

Each cell receives input from the previous cells in the same layer as well as the previous layer. After processing the inputs, the cell generates an output and propagates it to the next cells. Figure 3 shows an LSTM cell which consists of four components. Cell state stores information over time. The input gate determines how much data should be entered into the memory cell. Forget gate indicates what part of data should be discarded from going into the memory cell. Finally, the output gate controls the output of the memory cell, determining which information from the previous time step should be kept or forgotten. This allows the network to selectively remember or forget information over time. Relationships between LSTM cell components are as follows:

$$f_t = \sigma(W_{xf} x_t + W_{hf} h_{t-1} + b_f) \quad (3)$$
$$i_t = \sigma(W_{xi} x_t + W_{hi} h_{t-1} + b_i) \quad (4)$$
$$o_t = \sigma(W_{xo} x_t + W_{ho} h_{t-1} + b_o) \quad (5)$$
$$c_t = f_t \odot c_{t-1} + i_t \odot \tanh(W_{xc} x_t + W_{hc} h_{t-1} + b_c) \quad (6)$$
$$h_t = o_t \odot \tanh(c_t) \quad (7)$$

where $\sigma$ is sigmoid function, and $\odot$ is element-wise product and $f_t$, $i_t$ and $o_t$ are gating vectors.

## I. EXPERIMENTAL RESULTS

This section studies the effect of LSTM architecture, type of information, number of frames and number of surrounding vehicles on the LC prediction.

### A. Evaluation Metrics

The performance of the LC prediction model can be assessed based on accuracy, precision, and recall metrics. These criteria are calculated by using false positive (FP), false negative (FN), true negative (TN), and true positive (TP) counts.

$$Accuracy = \frac{TP+TN}{TP+TN+FP+FN} \quad (8)$$
$$Precision = \frac{TP}{TP+FP} \quad (9)$$
$$Recall = \frac{TP}{TP+FN} \quad (10)$$

TABLE II. LSTM PARAMETERS.

| Parameter | Description | Value |
|---|---|---|
| Input Dimension | CACC(8 surrounding cars, 5 frames, dp, dv, da) | 120 |
| | ACC(3 preceding cars, 5 frames, dp, dv) | 30 |
| Output Dimension | Dimension of output layer | 1 |
| Batch Size | Number of training cases over each optimizer update | 32 |
| Hidden Layer Number | Number of LSTM layers | 2 |
| Dropout rate | The rate used in dropout layers | 0.2 |
| Number of epochs | Number of training update | 100 |
| Loss function | Function to calculate loss | Binary cross entropy |
| Activation Function | Function to activate output of LSTM layers | Relu |
| Optimizer | The fuction to minimize loss | RMSprop |

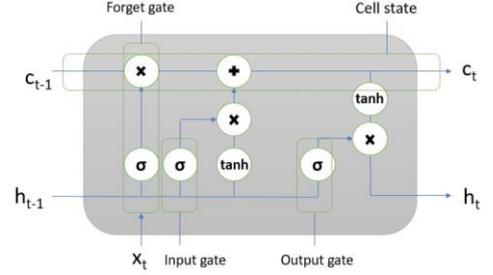

Figure 3. An LSTM cell.

### B. LSTM Architecture

In this experiment, we investigate the effect of LSTM cells number on the LC prediction. Table 3 shows that increasing the number of cells from 8 to 128 leads to an improvement in the LSTM performance, achieving an average increase of 2% in accuracy across all tested datasets. However, when we used 256 cells, the accuracy declined to 91.97 due to the overfitting problem, when multi-LSTM starts memorizing data. Therefore, 128 cells were selected for the next experiments.

### C. Vehicles' Information

To assess the influence of incorporating diverse information regarding surrounding vehicles on our metrics, we conducted an analysis of three distinct scenarios: solely *dp*, both *dp* and *dv*, and the complete set of available information encompassing *dp*, *dv*, and *da*. As illustrated in Table 4, augmenting the amount of information improves the LC accuracy. Nonetheless, the inclusion of additional data results in a significant surge in our execution time, as evidenced by the increase from 540s for solely *dp* to 986s and 1624s for *dp*, *dv*, and *dp*, *dv*, *da*, respectively.

TABLE III. EFFECT OF LSTM CELL NUMBERS.

| LSTM Units | Accuracy | Precision | Recall |
|---|---|---|---|
| 8 | 90.47 | 95.33 | 85.43 |
| 16 | 91.49 | 96.33 | 86.03 |
| 32 | 91.74 | 96.11 | 87.23 |
| 64 | 91.15 | 95.98 | 85.79 |
| 128 | 92.43 | 96.87 | 87.96 |
| 256 | 91.97 | 96.21 | 87.14 |

TABLE IV. EFFECT OF VEHICLE'S INFORMATION.

| Vehicle's Information | Accuracy | Precision | Recall |
|---|---|---|---|
| dp | 89.45 | 96.12 | 82.67 |
| dp, dv | 90.94 | 93.57 | 87.92 |
| dp, dv, da | 92.43 | 96.87 | 87.96 |

*D. Frame Set Size*

Time length (frame set size) before lane change that determines the input sequence length has a significant effect on the accuracy of predictions made by an LSTM network. If the frame set size is too short, the network may not have enough information to make accurate predictions. On the other hand, if the frame set size is too long, the network may suffer from irrelevant infromation, high computation, and vanishing gradient problem. Figure 4 shows that frame set of size 5 produces the best results for LC prediction.

*E. ACC and CACC Systems*

The number and location of vehicles with respect to the ego vehicle, are critical factors in the design of our system. In this regard, we present a comprehensive analysis of the performance metrics associated with four different scenarios shown in Table 5. Experiments demonstrate that the incorporation of information regarding the alongside vehicles fails to enhance the accuracy of the network, given that the presence of a vehicle in this region typically results in a LK. Furthermore, a comparison between the first and the last row of Table 5 reveals a significant improvement in all the metrics when utilizing all surrounding vehicles, in comparison to the scenario of utilizing solely three preceding vehicles.

## VI. CONCLUSION

This study explores the impact of vehicle-to-vehicle communication and the type of vehicles' information on the prediction accuracy of lane change. Results demonstrate that the proposed model, which employs an LSTM trained on the HighD dataset, achieves significant improvement in accuracy with an increase in the number of surrounding vehicles and their information. By changing our scenario from ACC to CACC, a 33.28% increase in accuracy was seen. Increasing the number of LSTM cells to 128 and selecting a frame set size of 5 leads to maximum accuracy. Additionally, using more information about other vehicles increases lane change prediction accuracy at the cost of a higher computation burden.

ACKNOWLEDGMENT

We acknowledge the financial support from the Natural Sciences and Engineering Research Council of Canada (NSERC) Catalyst Grant.

TABLE V. EFFECTS OF TYPE OF VEHICLES

| Type of Vehicles | Accuracy | Precision | Recall |
|---|---|---|---|
| LP, PV, RP | 59.15 | 78.54 | 28.43 |
| LF, FV, RF | 92.26 | 96.18 | 87.65 |
| LP, PV, RP, LA, RA | 62.89 | 75.29 | 34.8 |
| LP, PV, RP, LA, RA, LF, FV, RF | 92.43 | 96.87 | 87.96 |

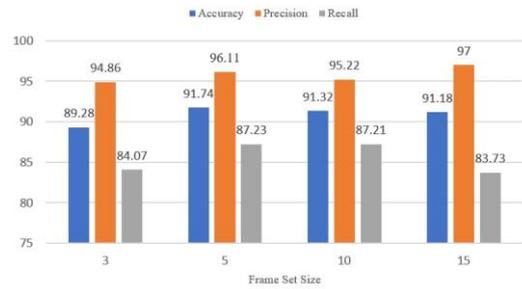

Figure 4. Effect of frame set size.